\documentclass[10pt,twocolumn,letterpaper]{article}

\usepackage{wacv}
\usepackage{times}
\usepackage{epsfig}
\usepackage{graphicx}
\usepackage{amsmath}
\usepackage{amssymb}

\usepackage{subcaption, booktabs}
\usepackage{xcolor, colortbl}
\usepackage{dblfloatfix}
\definecolor{gray}{rgb}{0.8,0.8,0.8}
\newcommand{\tablecolor}{\cellcolor{gray}}
\colorlet{RED}{red}
\newcommand\sectionprelude{%
  \vspace{14em}
}
\graphicspath{ {./figures/} }

\def\wacvPaperID{****} 

\wacvfinalcopy 

\ifwacvfinal
\def\assignedStartPage{1} 
\fi

\ifwacvfinal
\usepackage[breaklinks=true,bookmarks=false]{hyperref}
\else
\usepackage[pagebackref=true,breaklinks=true,colorlinks,bookmarks=false]{hyperref}
\fi

\ifwacvfinal
\else
\fi

\begin{document}

\title{Improving Building Segmentation for Off-Nadir Satellite Imagery}

\author{\parbox{16cm}{\centering
    {Hanxiang Hao$^{\star}$ \quad Sriram Baireddy$^{\star}$ \quad Kevin LaTourette$^{\dagger}$ \quad Latisha Konz$^{\dagger}$ \quad Moses Chan$^{\dagger}$ Mary L. Comer$^{\star}$ \quad Edward J. Delp$^{\star}$}\\
    {\normalsize
    $^{\star}$ Video and Image Processing Lab (VIPER), Purdue University, West Lafayette, Indiana USA\\
    $^{\dagger}$ Lockheed Martin Space,
    Sunnyvale, California USA}}
}

\twocolumn[{%
\renewcommand\twocolumn[1][]{#1}%
\maketitle
\begin{center}
    \centering
    \captionsetup{type=figure}
    \includegraphics[width=1\textwidth]{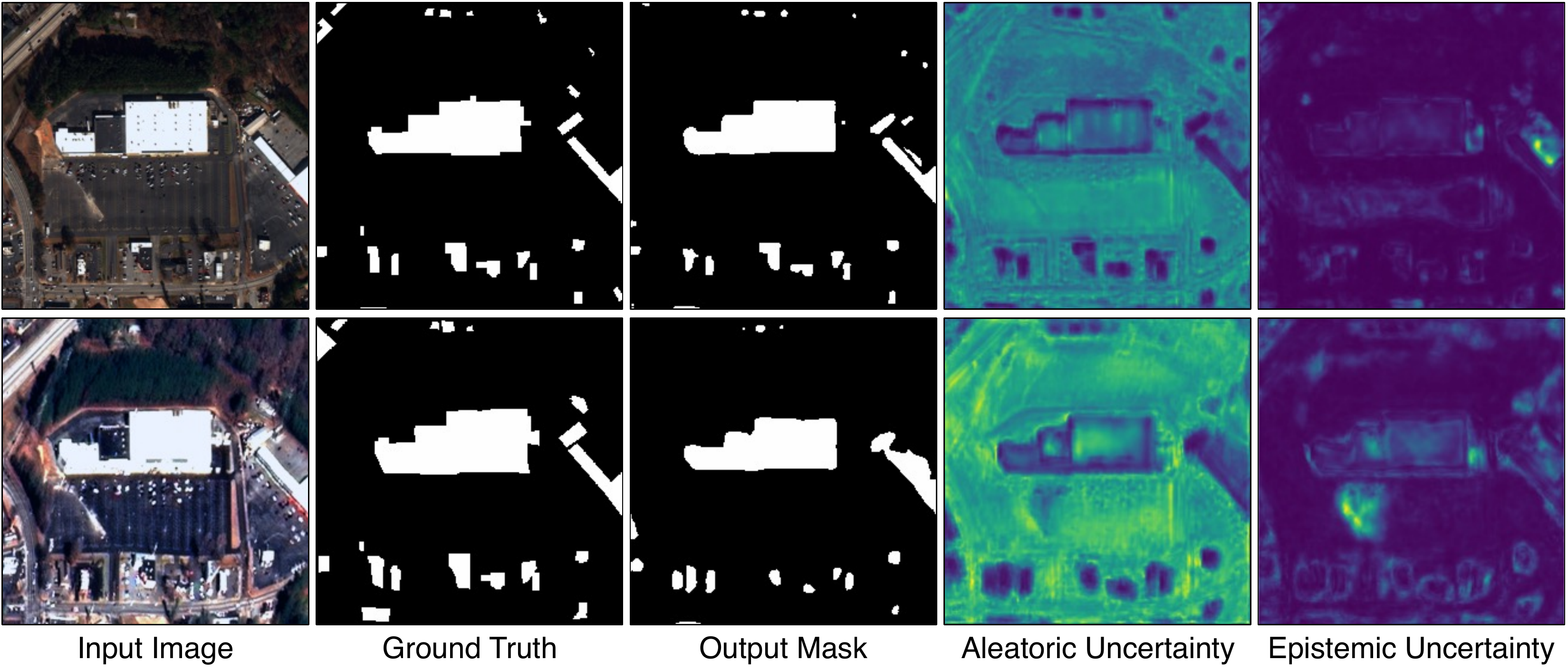}
    \captionof{figure}{Building segmentation results of the proposed method with corresponding uncertainty maps. The first row shows the case with off-nadir angle as $-7.8^\circ$. The second row shows the result of the same scene but with off-nadir angle as $54^\circ$. With a larger off-nadir angle, the input image becomes noisy and blurry. Since aleatoric uncertainty captures the noise inherent in the observations, higher values can be found in the aleatoric uncertainty map from the second case. Class-ambiguous pixels are highlighted in the epistemic uncertainty maps, which often yield incorrect classification results.}
    \label{fig:teaser}
\end{center}%
}]

\maketitle

\begin{abstract}
  Automatic building segmentation is an important task for satellite imagery analysis and scene understanding.
  Most existing segmentation methods focus on the case where the images are taken from directly overhead (\ie low off-nadir/viewing angle). 
  These methods often fail to provide accurate results on satellite images with larger off-nadir angles due to the higher noise level and lower spatial resolution.
  In this paper, we propose a method that is able to provide accurate building segmentation for satellite imagery captured from a large range of off-nadir angles.
  Based on Bayesian deep learning, we explicitly design our method to learn the data noise via aleatoric and epistemic uncertainty modeling. 
  Satellite image metadata (\eg off-nadir angle and ground sample distance) is also used in our model to further improve the result.
  We show that with uncertainty modeling and metadata injection, our method achieves better performance than the baseline method, especially for noisy images taken from large off-nadir angles.\footnote{This is an extended version of our SIGSPATIAL'21 paper~\cite{Hao_2021b}}
\end{abstract}

\section{Introduction}

Object segmentation for satellite imagery has been studied extensively because of the availability of large-scale datasets~\cite{Etten_2018, Weir_2019, Gupta_2019, Chiu_2020, Etten_2021} and computational resources.
Although many existing methods achieve accurate segmentation results, using them in real-world applications is still challenging.
Unlike many segmentation tasks for natural images, such as the COCO dataset~\cite{Lin_2014} and Cityscapes dataset~\cite{Cordts_2016}, real-world object segmentation for satellite imagery often faces challenges in identifying small, visually heterogeneous objects (\eg cars and buildings) with varying orientation and density in images~\cite{Weir_2019}.
For example, it is even hard for humans to detect the small buildings inside the forest area from the images in Figure~\ref{fig:teaser}, because of the low lighting condition and the similar colors of the buildings compared to their surrounded trees. 
Furthermore, due to changes in the satellite viewing angle, the appearance of target objects can vary dramatically, including changes in lighting intensity, object resolution, and image noise level. 
As the input images in Figure~\ref{fig:teaser} show, from small viewing angle (first row) to large viewing angle (second row), the overall image intensity and image quality changes significantly.
Therefore, to be able to successfully use the segmentation models in real-world applications, addressing the aforementioned challenges is necessary. 

\begin{figure}[t]
  \centering
  \includegraphics[width=0.65\linewidth]{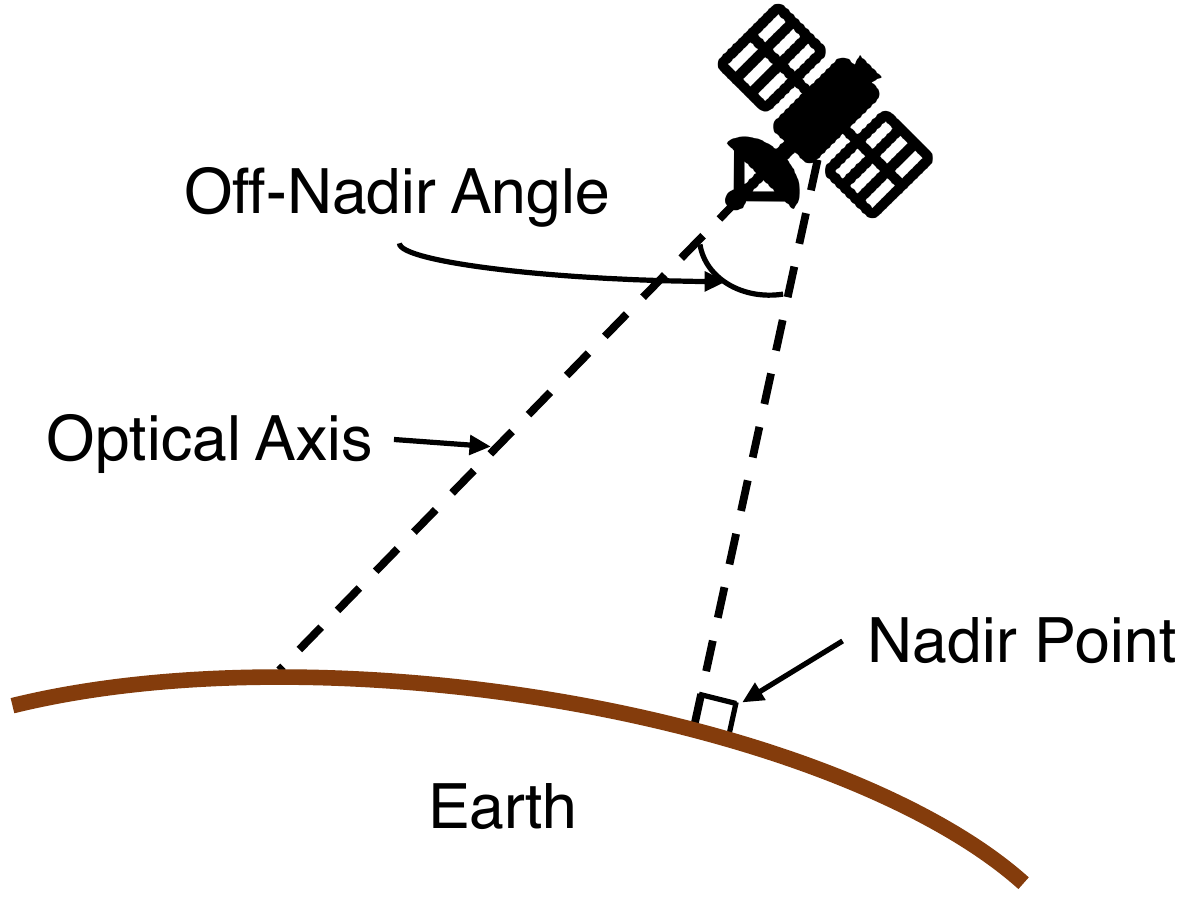}
  \caption{Illustration of satellite off-nadir angle.}
  \label{fig:off-nadir}
\end{figure}

Many existing satellite imagery segmentation methods directly adopt approaches that were originally designed for the natural image object segmentation task without considering the previously mentioned challenges.
Since most of the publicly available datasets for satellite image segmentation consist of images taken nearly directly overhead (\ie at-nadir images)~\cite{Etten_2018, Gupta_2019, Chiu_2020, Etten_2021}, these existing methods are able to produce accurate results.
However, as mentioned earlier, the accurate results do not guarantee that these methods can be successfully used in real-world applications.
To address this issue, in this work, we consider the more challenging SpaceNet 4, a multi-view overhead imagery dataset~\cite{Weir_2019} for building segmentation, which focuses on noisy data due to large off-nadir angles.
As shown in Figure~\ref{fig:off-nadir}, satellite off-nadir angle (\ie viewing angle) is the angle between the nadir point directly below the satellite and the center of the imaged scene~\cite{Weir_2019}.
Considering images with large off-nadir angles enables us to move one step closer to real-world applications.
For example, many satellite images collected during disaster responses or other urgent situations often involve large off-nadir angles.
The first set of satellite images taken from Puerto Rico after Hurricane Maria was obtained with the off-nadir angle as $51.9^\circ$~\cite{DigitalGlobe}. 
A large off-nadir angle can cause a significant deterioration in image quality. 
As shown in Figure~\ref{fig:teaser}, compared to the image with the smaller off-nadir angle, the image with the larger off-nadir angle is blurrier and noisier. 
Furthermore, a large off-nadir angle can also cause a change in object appearance. 
For example, in the same figure, with the smaller off-nadir angle, only building roofs are visible, but with the larger off-nadir angle, both building roofs and their facades are visible, which will cause the change of building area in the satellite images.
In the SpaceNet 4 dataset, images of the same scene are taken at different off-nadir angles.
All building annotations are labeled based on the images with the smallest magnitude of off-nadir angle ($-7.8^\circ$) and the rest of the images with different off-nadir angles use the same labels as ground truth during training.
Therefore, the change of building appearance due to the change of off-nadir angle has an adverse effect when training the model due to the inaccurate ground truth annotations.
These challenges are similar to the challenges in domain adaptation, where reliable data is available for training in one scenario, but the model needs to be adapted to new data collected under different scenarios (\eg different lighting conditions, image noise conditions, or annotation accuracy conditions).

In order to solve these challenges provided by the SpaceNet 4 dataset and real-world applications, we present a building segmentation method with uncertainty modeling and satellite image metadata injection.
Our method is able to provide accurate segmentation results when training with noisy images and inaccurate ground truth annotations.
More specifically, based on Bayesian deep learning, the proposed method is designed to capture both model and data uncertainty to ignore the image regions with a higher uncertainty level.
For example, as shown in Figure~\ref{fig:teaser} (we will provide more detailed information in Section~\ref{sec:uncertainty}), our uncertainty maps highlight the areas with larger image noise (\eg building boundaries due to the image blur and inaccurate annotation). 
As the off-nadir angle increases (\ie from the first row to the second row), the uncertainty level increases, indicating a higher data noise from both image and annotation. 
Furthermore, satellite image metadata is also considered in our method, as it usually contains useful information to improve model performance.
In this work, we use ground sample distance (GSD) and off-nadir angle as input metadata.
GSD describes the spatial resolution of the image and a larger GSD usually indicates blurrier and noisier images.
As mentioned earlier, different off-nadir angles can also cause changes in image quality.
In this work, we propose two metadata injection methods in Section~\ref{sec:metadata} to show the effectiveness of using metadata in building segmentation. 
The main contributions of this paper are summarized as follows:
\begin{itemize}
  \item we design a building segmentation model that is able to capture both model uncertainty (\ie epistemic uncertainty) and data uncertainty (\ie aleatoric uncertainty);
  \item a concatenation-based metadata injection method is developed for using satellite image metadata to improve building segmentation;
  \item alternatively, we also propose a metadata injection method using Affine Combination Module for multi-resolution injection;
  \item based on our experimental analysis and ablation study, we show that the proposed method is able to achieve a better performance than the baseline method, especially for noisy images taken at large off-nadir angles. 
\end{itemize}

\section{Related Work}

In this section, we will review the previous work for satellite image building segmentation as well as the methods using uncertainty modeling and metadata injection in satellite imagery. 

\textbf{Building segmentation for satellite imagery.}
In this paper, we consider the building segmentation task as a binary semantic segmentation task\footnote{Some previous work also considered this task as an instance segmentation task. In this paper, we will focus on the semantic segmentation task.}.
Many recent approaches (including our proposed method) are designed based on the U-Net structure~\cite{Ronneberger_2015}, because of its good performance in many computer vision tasks~\cite{Lin_2017, Badrinarayanan_2017, Isola_2017, Iglovikov_2018, Chen_2018}.
Here we briefly review several U-Net-based methods of building segmentation for satellite imagery. 
A large receptive field is important for the segmentation model to detect buildings with different sizes. 
Therefore, many methods improve the original U-Net by using different techniques to enlarge the receptive field to achieve better performance.
Zhang~\etal~\cite{Zhang_2019} extend the U-Net model with dense connections~\cite{Huang_2017} and dilated convolutional layers~\cite{Yu_2016, Chen_2018} to reach a large receptive field for capturing the information of large objects.
Liu~\etal~\cite{Liu_2021} incorporate a feature pyramid scene parsing (PSP) network~\cite{Zhao_2017} with U-Net to further increasing the receptive field.
They use the PSP module to replace the bottleneck layer from U-Net to allow the use of multi-scale features for extracting building footprints of different sizes.
Jing~\etal~\cite{Jing_2021} design a spatial pyramid dilated network for building segmentation by combining the aforementioned PSP network with dilated convolution.
In this work, as discussed previously, instead of focusing on improving the performance on the at-nadir images, our method aims to deal with the problem of adapting for real-world applications: building segmentation for images with large off-nadir angles, as these images tend to be noisier and blurrier than at-nadir images. 

\textbf{Uncertainty modeling for satellite imagery analysis.} 
Using Bayesian deep learning to model uncertainty has already been seen in satellite imagery analysis.
Kampffmeyer~\etal~\cite{Kampffmeyer_2016} first introduced Monte Carlo dropout~\cite{Gal_2016a} to capture model uncertainty for small object segmentation.
Although dropout is rarely used in convolutional neural networks (CNNs) due to the empirically deteriorated performance, they show that adding dropout layers in their fully convolutional encoder-decoder model with Monte Carlo integration during inference can achieve better performance.
Our proposed method also uses Monte Carlo dropout; please check Section~\ref{sec:epistemic} for more information.
Inspired by this, Bischke~\etal~\cite{Bischke_2018} proposed to use the model uncertainty to address the class imbalance issue in the satellite image segmentation task.
The predicted uncertainty for each class is used as the weight in the cross-entropy loss to account for model uncertainty caused by class imbalance.
In this work, we propose to use not only the model uncertainty (\ie epistemic uncertainty) as presented in the previous work, but also the data uncertainty (\ie aleatoric uncertainty) to enable our segmentation model to learn from noisy data.

\textbf{Injecting metadata for satellite imagery analysis.}
Satellite image metadata can be used in many satellite imagery analysis tasks, as it usually contains useful information to improve model performance.
Pritt~\etal~\cite{Pritt_2017} use a variety of satellite metadata including GSD, off-nadir angle, longitude, and latitude for the image classification task in satellite imagery.
They use an ensemble of CNN models for image feature extraction. 
Then the CNN features are concatenated with the normalized metadata and fed into fully-connected layers for classification.
In Section~\ref{sec:metacat}, we will provide a similar concatenation-based metadata injection method with an improvement of metadata feature extraction using multi-layer perceptrons.
Christie~\etal~\cite{Christie_2018} proposed a similar model to fuse the CNN features with normalized metadata for multi-temporal satellite image sequence.
Different from the previous work, instead of feeding the fused features to fully-connected layers, these features are fed into a long short-term memory (LSTM) model to accumulate temporal information from different frames to obtain the final classification result.
In this work, besides the aforementioned concatenation-based method, we will also present an Affine Combination Module-based metadata injection to inject metadata for multiple feature resolutions as described in Section~\ref{sec:metaacm}. 

\section{Method}

\begin{figure*}[t]
  \centering
  \includegraphics[width=0.7\linewidth]{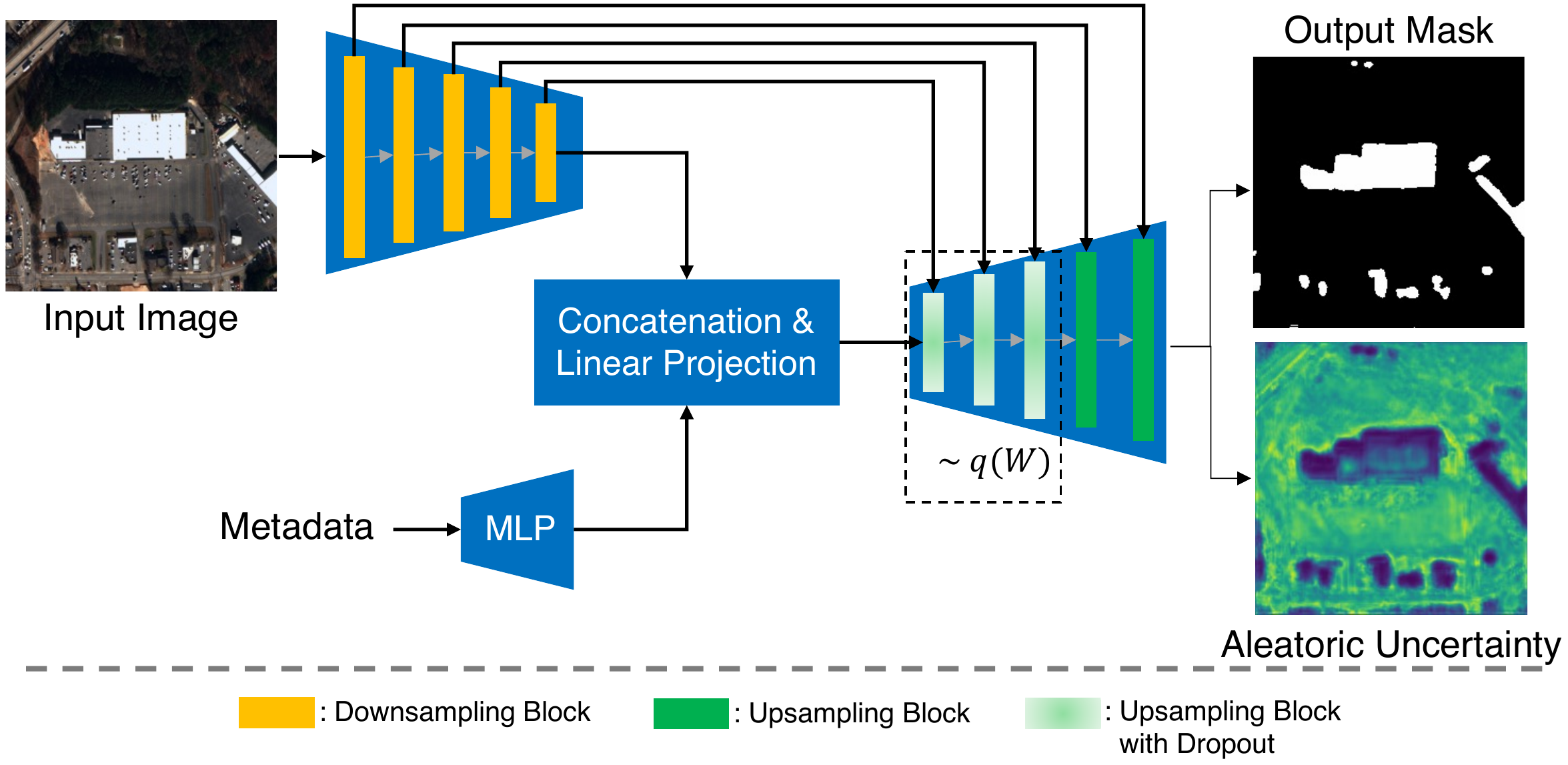}
  \caption{The block diagram of the proposed method with uncertainty modeling and concatenation-based metadata injection. 
  $q(W)$ is the dropout variational distribution as described in Section~\ref{sec:epistemic}.}
  \label{fig:method_cat}
\end{figure*}

In this section, we will introduce our building segmentation method with uncertainty modeling and satellite image metadata injection.
As shown in Figure~\ref{fig:method_cat}, the proposed method is based on U-Net~\cite{Ronneberger_2015} and has multiple outputs.
As described in Section~\ref{sec:uncertainty}, modeling uncertainty enables our method to ignore the noisy pixels that are caused by 1) blurry or noisy images; and 2) inaccurate data annotation. 
Injecting satellite image metadata such as ground sample distance (GSD) and off-nadir angle provides the model with more information to improve its performance. 
We will provide two metadata injection approaches in Section~\ref{sec:metadata}.

\subsection{Modeling Uncertainty via Bayesian Deep Learning}
\label{sec:uncertainty}

Unlike standard deep learning methods, Bayesian deep learning (Bayesian DL) provides a model with the ability to ignore certain data points based on uncertainty.
In Bayesian DL, there are two types of uncertainty one can model:
\begin{itemize}
  \item \textit{Epistemic Uncertainty} describes the uncertainty that is caused by the model ignoring some training data. 
  For example, a segmentation model might miss some building areas with certain colors/textures. 
  Usually, this type of uncertainty can be reduced as more training data is made available. 
  \item \textit{Aleatoric Uncertainty} describes the uncertainty that is inherited from data (\eg image/sensor noise). 
  Aleatoric uncertainty can be further categorized as \textit{homoscedastic uncertainty}, which is the uncertainty based on the entire dataset, and \textit{heteroscedastic uncertainty}, which is the uncertainty for each input data point (\ie each pixel in our case). 
  In this work, we will consider heteroscedastic aleatoric uncertainty to accurately model the data noise for different input images. 
\end{itemize}
In the following section, we will review the methods for modeling epistemic uncertainty~\cite{Gal_2016b} and aleatoric uncertainty~\cite{Kendall_2017a}, followed by our proposed approach to combine both uncertainties in one model. 

\subsubsection{Epistemic Uncertainty}
\label{sec:epistemic}

In Bayesian DL, to capture the uncertainty from the model (\ie epistemic uncertainty), we place a distribution over the model parameters.
For example, the prior distribution of the model weights for a fully-connected layer, $p(\mathbf{W})$, can be modeled as: $\mathbf{W} \sim \mathcal{N}(\mathbf{0}, \mathbf{I})$.
This is different from the standard deep learning model, which uses deterministic parameters. 
In Bayesian DL, for each forward pass, including both training and testing, the model parameters will be different due to parameter sampling.
Formally speaking, we formulate our building segmentation model as:
\begin{equation} \label{eq:inference}
  p(\mathbf{y} | \mathbf{x}, \mathbf{W}) = p(\mathbf{y} | f^{\mathbf{W}}(\mathbf{x})) = S(f^{\mathbf{W}}(\mathbf{x})),
\end{equation}
where $\mathbf{x} \in \mathbb{R}^{H \times W \times C}$ is the input image, $\mathbf{y}  \in \mathbb{R}^{H \times W}$ is the output class label (in our case, it is a binary label indicating foreground or background), $f^{\mathbf{W}}: \mathbb{R}^{H \times W \times C} \rightarrow \mathbb{R}^{H \times W}$ is our Bayesian DL model with sampled parameters $\mathbf{W} \sim p(\mathbf{W} | \mathbf{X}, \mathbf{Y})$, and $S: \mathbb{R} \rightarrow \mathbb{R}$ is the sigmoid function applied to each input element.

Estimating the model posterior $p(\mathbf{W} | \mathbf{X}, \mathbf{Y})$ over the entire training set $(\mathbf{X}, \mathbf{Y})$ is intractable~\cite{Gal_2016a, Gal_2016b}. 
To evaluate this posterior distribution, following the work~\cite{Gal_2016a, Gal_2016b, Kendall_2017a, Kendall_2017b}, we use \textit{dropout variational inference}.
This inference is performed by placing a dropout layer before every convolutional layer (or fully-connected layer).
Since dropout can be formulated as a Bernoulli trial by randomly setting the model parameters to zero, \cite{Gal_2016a, Gal_2016b} show that this dropout distribution over model parameters, $q(\mathbf{W})$, can be used to estimate our model posterior. 
This is done by minimizing their Kullback-Leibler (KL) divergence via the following loss function during training:
\begin{equation} \label{eq:epistemic}
  \mathcal{L}_{epi}(\mathbf{x}, \mathbf{y}) = \mathcal{L}_{cls}(\mathbf{y}, S(f^{\mathbf{W}}(\mathbf{x}))) + \lambda \Vert \mathbf{W} \Vert_2^2, 
\end{equation}
where $(\mathbf{x}, \mathbf{y})$ is a pair of training image and its corresponding ground truth label mask, $\mathcal{L}_{cls}(\cdot, \cdot)$ is a classification loss (\eg binary cross entropy loss in our case), $f^{\mathbf{W}}(\cdot)$ is our model with parameters sampled from the dropout distribution $q(\mathbf{W})$, and $\lambda$ is a non-trainable hyper-parameter as described in~\cite{Gal_2016a}. 
The second term of Equation~\ref{eq:epistemic} can be implemented using weight decay~\cite{Krogh_1991}, which was originally designed for model regularization. 
During inference, we can estimate the final prediction distribution $p(\textbf{y}^* | \textbf{x}^*)$ given a testing image $\mathbf{x}^*$ via Monte Carlo integration as proposed in~\cite{Gal_2016a, Gal_2016b}: 
\begin{equation} \label{eq:mc_dropout}
  p(\textbf{y}^* | \textbf{x}^*) = \int p(\textbf{y}^* | \textbf{x}^*, \textbf{W})q(\textbf{W}) d\textbf{W} \approx S(\frac{1}{T} \sum_{t=1}^T f^{\mathbf{W}_t}(\mathbf{x}^*)),
\end{equation}
where $\mathbf{W}_t \sim q(\mathbf{W})$ is the model parameters from each Monte Carlo sample and $T$ is the total number of samples. 
Equation~\ref{eq:mc_dropout} is referred as Monte Carlo dropout as proposed in~\cite{Gal_2016a}.
Epistemic uncertainty can be visualized by calculating the variance of the Monte Carlo samples:
\begin{equation}
  \frac{1}{T} \sum_{t=1}^T \left( f^{\mathbf{W}_t}(\mathbf{x}) \odot f^{\mathbf{W}_t}(\mathbf{x}) \right) - \mathbb{E}[f^{\mathbf{W}}(\mathbf{x})] \odot \mathbb{E}[f^{\mathbf{W}}(\mathbf{x})], 
\end{equation}
where $\odot$ is the Hadamard product for element-wise multiplication and $\mathbb{E}[f^{\mathbf{W}}(\mathbf{x})] \approx \frac{1}{T} \sum_{t=1}^T f^{\mathbf{W}_t}(\mathbf{x})$.

As shown in Figure~\ref{fig:method_cat}, we model the epistemic uncertainty by placing the dropout layers before just the first three decoder layers, instead of all convolutional layers. 
Since we use a ResNet-34 model~\cite{He_2016} pretrained on ImageNet~\cite{Deng_2009} as the CNN encoder, we model this feature extraction process as a deterministic process. 
Therefore, no dropout layers are used in the CNN encoder.
In this work, we only model the first three decoder layers as stochastic processes by placing the dropout layers before each convolutional layer in each upsampling block.
We do not add dropout layers to the last two decoder layers. 
This is to reduce the output noise due to the limited number of Monte Carlo samples during inference as shown in Equation~\ref{eq:mc_dropout}.

\subsubsection{Aleatoric Uncertainty}
\label{sec:aleatoric}

Aleatoric uncertainty captures the noise from training data.
As described previously, in this work, we consider heteroscedastic aleatoric uncertainty, which captures the noise from each pixel from an input image.
We use two additional convolutional layers placed on top of the last decoder layer to obtain the classification logit $f^{\mathbf{W}}(\mathbf{x}) \in \mathbb{R}^{H \times W}$ and aleatoric uncertainty $\sigma^{\mathbf{W}}(\mathbf{x}) \in \mathbb{R}^{H \times W}$, as shown in Figure~\ref{fig:method_cat}.
We use the predicted aleatoric uncertainty during training to ignore the pixels with larger uncertainty and address the pixels with less uncertainty.
To achieve this, as proposed in~\cite{Kendall_2017a}, we corrupt the predicted logits $f^{\mathbf{W}}(\mathbf{x})$ with Gaussian random noise, where the standard deviation is the predicted aleatoric uncertainty. 
More specifically, we modify Equation~\ref{eq:inference} by placing a Gaussian distribution over the predicted logits:
\begin{equation} \label{eq:inference_aleatoric}
\begin{aligned} 
  & p(\mathbf{y}_{i, j} | \mathbf{x}, \mathbf{W}) = S(\hat{f}^{\mathbf{W}}(\mathbf{x})_{i,j}), \\
  \text{where } & \hat{f}^{\mathbf{W}}(\mathbf{x})_{i,j} \sim \mathcal{N}(f^{\mathbf{W}}(\mathbf{x})_{i,j}, (\sigma^{\mathbf{W}}(\mathbf{x})_{i,j})^2).
\end{aligned}
\end{equation}
Note that $i$ and $j$ are the pixel coordinates of the output logit and aleatoric uncertainty. 
We denote $p(\mathbf{y}_{i, j} | \mathbf{x}, \mathbf{W})$ with $p_{i,j}$ for simplicity.
From Equation~\ref{eq:inference_aleatoric}, we can see that with larger aleatoric uncertainty, the Gaussian corrupted logit $\hat{f}^{\mathbf{W}}(\mathbf{x})$ tends to be noisier, which enforces the model to ignore this ``random'' prediction. 
With smaller aleatoric uncertainty, the Gaussian corrupted logit $\hat{f}^{\mathbf{W}}(\mathbf{x})$ tends to be closer to the original predicted logit $f^{\mathbf{W}}(\mathbf{x})$, which makes the model to focus on this prediction.
Since we use Gaussian corruption, we can facilitate our implementation using the Gaussian reparameterization trick:
\begin{equation} \label{eq:reparameterization}
  \hat{f}^{\mathbf{W}}(\mathbf{x})_{i,j} = f^{\mathbf{W}}(\mathbf{x})_{i,j} + \sigma^{\mathbf{W}}(\mathbf{x})_{i,j} \epsilon, \ \ \ \epsilon \sim \mathcal{N}(0, 1).
\end{equation} 
During training, to capture both uncertainties, we can replace the classification loss $\mathcal{L}_{cls}$ in Equation~\ref{eq:epistemic} with a binary cross entropy loss with Gaussian corrupted output:
\begin{equation} \label{eq:aleatoric}
  \mathcal{L}_{ale}(\mathbf{x}, \mathbf{y}) = -\frac{1}{HW} \sum_{i=1}^H \sum_{j=1}^W \mathbf{y}_{i,j} \log{p_{i,j}} + (1 - \mathbf{y}_{i,j}) \log{(1 - p_{i,j}}),
\end{equation}
where $\mathbf{y}_{i,j}$ is ground truth label at pixel coordinates $(i, j)$ and $p_{i,j} = S(\hat{f}^{\mathbf{W}}(\mathbf{x})_{i,j})$ as shown in Equation~\ref{eq:inference_aleatoric}.
Therefore, we can obtain the final loss function for learning both epistemic uncertainty and aleatoric uncertainty as:
\begin{equation} \label{eq:both}
  \mathcal{L}_{both}(\mathbf{x}, \mathbf{y}) = \mathcal{L}_{ale}(\mathbf{x}, \mathbf{y})  + \lambda \Vert \mathbf{W} \Vert_2^2.
\end{equation}
We do not need aleatoric uncertainty during inference, as it is used for ignoring noisy pixels during training.

\subsection{Metadata Injection}
\label{sec:metadata}

Satellite image metadata contains useful information to support many computer vision tasks, such as using solar and satellite azimuth and elevation angles for shadow detection and building height estimation~\cite{Raju_2014, Liasis_2016, Gouiaa_2018, Trekin_2020, Hao_2021}. 
In this paper, we consider two types of metadata to improve the building segmentation result: (1) ground sample distance (GSD); and (2) off-nadir angle. 
GSD describes the spatial resolution of the image; a larger GSD indicates blurrier and noisier images due to lower image resolution.
Off-nadir angle describes the viewing angle of the satellite camera and a larger off-nadir angle can also cause lower image resolution.
In the following sections, we will provide two metadata injection approaches to improve the baseline U-Net model.

\subsubsection{Metadata Injection via Feature Concatenation}
\label{sec:metacat}

As shown in Figure~\ref{fig:method_cat}, we first pass the metadata vector to multi-layer perceptrons (MLP) to obtain the output vector ($\mathbf{h} \in \mathbb{R}^D$) for feature extraction and dimension expansion.
Then we combine the metadata feature vector with the image features ($\mathbf{v} \in \mathbb{R}^{H \times W \times D}$) obtained from the last CNN encoder layer.
To combine metadata and image features, we repeat the metadata feature vector to match the shape of image features, getting $\mathbf{h}' \in \mathbb{R}^{H \times W \times D}$.
Then we concatenate the features along the channel dimension as $\mathbf{h}_{\mathbf{v}} \in \mathbb{R}^{H \times W \times 2D}$.
The final features can be obtained by linearly projecting the channel dimension back to the input channel dimension: $\mathbf{o} = \mathbf{F}(\mathbf{h}_{\mathbf{v}}) \in  \mathbb{R}^{H \times W \times D}$, where $\mathbf{F}: \mathbb{R}^{2D} \rightarrow \mathbb{R}^{D}$ is applied for each input element and it can be implemented by a convolutional layer with kernel size of $1$.
We refer to this concatenation-based approach as \textit{MetaCat}.

\subsubsection{Metadata Injection via Affine Combination Module}
\label{sec:metaacm}

\begin{figure*}[htbp]
  \centering
  \includegraphics[width=0.7\linewidth]{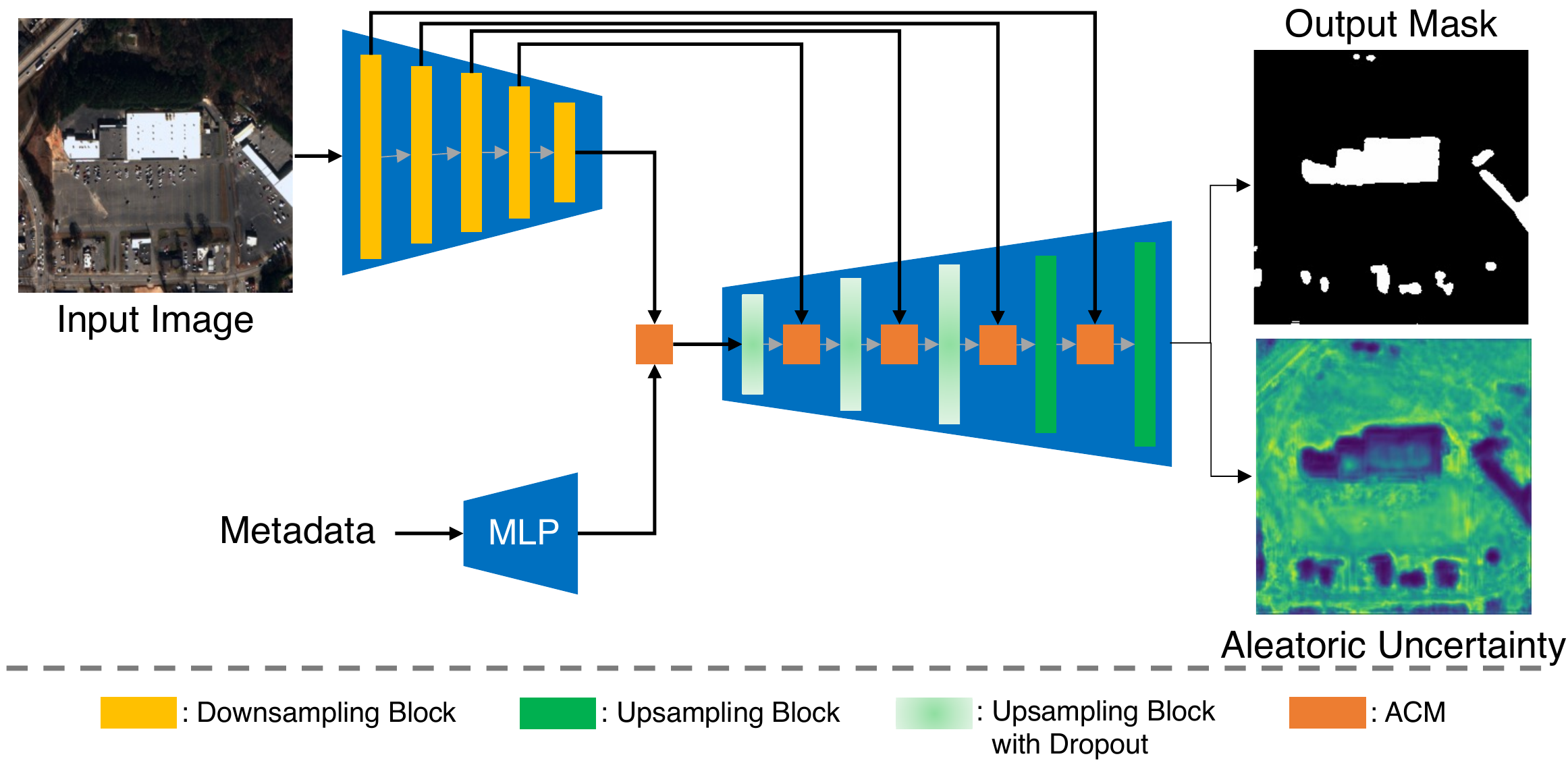}
  \caption{The block diagram of the proposed method with uncertainty modeling and ACM-based metadata injection.}
  \label{fig:method_acm}
\end{figure*}

As described above, the previous concatenation-based metadata injection method combines the metadata and image features by channel-wise concatenation following a linear projection layer.
By doing so, we augment the image features using the metadata features for every location in the $H$ and $W$ dimensions evenly.
However, intuitively, not all image features need to be modified.
For example, since we focus on building segmentation, a large forest area should not be considered and modified.
To effectively locate the desired regions that need to be modified, we use the Affine Combination Module (ACM)~\cite{Li_2020} for metadata injection as shown in Figure~\ref{fig:method_acm}.
As the name indicates, ACM is based on affine transforms and can be formulated as follows:
\begin{equation} \label{eq:acm}
  \mathbf{v}' = \mathbf{h} \odot W(\mathbf{v}) + b(\mathbf{v}), 
\end{equation}
where $\mathbf{v}$ is the image features obtained from the CNN encoder, $\mathbf{h}$ is either the repeated metadata features $\mathbf{h}'$ as described in Section~\ref{sec:metacat} or the features from the previous decoder layer, and $W(\cdot)$ and $b(\cdot)$ are convolutional layers as proposed in~\cite{Li_2020}. 
From Equation~\ref{eq:acm}, we can consider the $W(\mathbf{v})$ term as the metadata-relevant information, since it can directly interact with the metadata features (or the previous decoder features). 
The $b(\mathbf{v})$ term can be considered as a metadata-irrelevant information that is not modified by the metadata features (or the previous decoder features).
As the results that we will provide in Section~\ref{sec:exp_result} indicate, with ACM, we can explicitly decouple the metadata-relevant and metadata-irrelevant information without implicit learning by the model.
Following the design from~\cite{Li_2020}, we use multiple ACMs in different feature resolutions in our decoder without changing other parts of the model, as shown in Figure~\ref{fig:method_acm}.
We refer to this ACM-based approach as \textit{MetaACM}.

\section{Experiment}
\label{sec:exp}

In this section, we will describe the dataset we used and the model implementation details in Section~\ref{sec:exp_setting} while providing experimental results with analysis in Section~\ref{sec:exp_result}. 

\subsection{Dataset and Experiment Setting}
\label{sec:exp_setting}

In this work, we use the SpaceNet 4 dataset~\cite{Weir_2019}, which is designed for building segmentation with a larger range of off-nadir angles.
It contains 4-channel RGB-NIR (Near-Infrared) images with resolutions of $900 \times 900$.
There are $1,064$ distinct locations in the dataset, with $27$ images captured at each location at off-nadir angles ranging from $-32.5^\circ$ to $54^\circ$, which totals to $28,728$ images.
We partition the dataset into training, validation, and testing sets with the ratio of $6:2:2$.
Note that when splitting the dataset, we ensure that all images of the same location are assigned to the same partition. 
This can avoid different partitions sharing images from the same location. 

\begin{figure}[htbp]
  \centering
  \includegraphics[width=\linewidth]{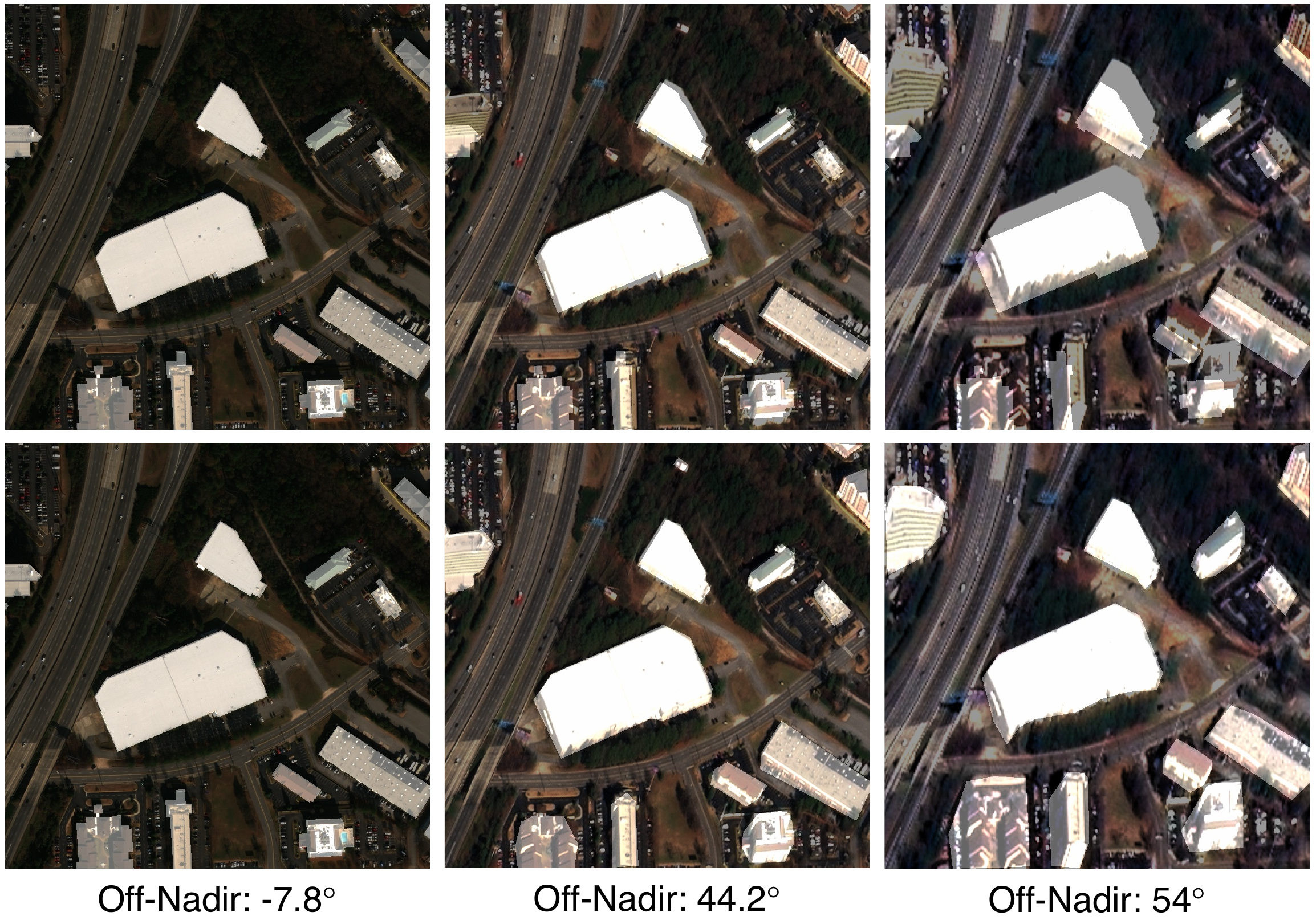}
  \caption{Illustration of the building segmentation annotation issue in the original dataset. The light white area is the annotated ground truth area. The first row shows the annotation from the original dataset and the second row shows the annotation we manually labeled.}
  \label{fig:dataset_issue}
\end{figure}

As mentioned in~\cite{Weir_2019}, the building annotations from the SpaceNet 4 dataset are obtained from the images with the smallest (in magnitude) off-nadir angle ($-7.8^\circ$), and the same annotations are used for the other images with different off-nadir angles.
As shown in the first row of Figure~\ref{fig:dataset_issue}, due to the change of viewing angle, the appearance, especially for the tall buildings, changes significantly.
For example, with the smaller off-nadir angle, only the building roof is visible, but with larger off-nadir angles, both building roof and facade are visible, which can cause inaccurate annotations.
Although the proposed method is designed to deal with the noisy images and annotations, in order to have an accurate testing evaluation, we manually label the testing images with off-nadir angles greater than $40^\circ$, as shown in the second row of Figure~\ref{fig:dataset_issue}.

To ensure fair comparison between the proposed method and the baseline U-Net, all of our experiments used the same setting, which we will now describe.
The downsampling blocks (yellow blocks) in Figure~\ref{fig:method_cat} and Figure~\ref{fig:method_acm} are the residual blocks from a ResNet-34 model~\cite{He_2016} pretrained on ImageNet~\cite{Deng_2009}.
The upsampling blocks (dark green blocks) consist of \textit{bilinear upsampling} $\rightarrow$ \textit{convolution} $\rightarrow$ \textit{batch normalization} $\rightarrow$ \textit{ReLU}. 
The upsampling blocks with dropout (light green blocks) consist of \textit{bilinear upsampling} $\rightarrow$ \textit{dropout} $\rightarrow$ \textit{convolution} $\rightarrow$ \textit{batch normalization} $\rightarrow$ \textit{ReLU}. 
Following~\cite{Kendall_2017a}, the dropout rate is set as $0.2$.
The MLP for metadata feature extraction consists of three blocks, where each block is a fully-connected layer following by a leaky ReLU layer with a slope of $0.2$. 
During training, to allow for a larger batch size as required by batch normalization, we resize the input image to $256$ with batch size as $64$.
ADAM optimizer~\cite{Kingma_2015} with learning rate $0.0001$ (linear decay) is used and all experiments are trained for 1 million iterations.
As mentioned in Section~\ref{sec:epistemic}, modeling epistemic uncertainty requires using weight decay during training.
To achieve a fair comparison, we use weight decay with the factor of $0.0001$ for all experiments.
For the Monte Carlo integration during inference, following~\cite{Kendall_2017a}, we set the number of samples as $50$ (we will provide the analysis of this parameter in the following section).

\begin{figure}[htbp]
  \centering
  \includegraphics[width=\linewidth]{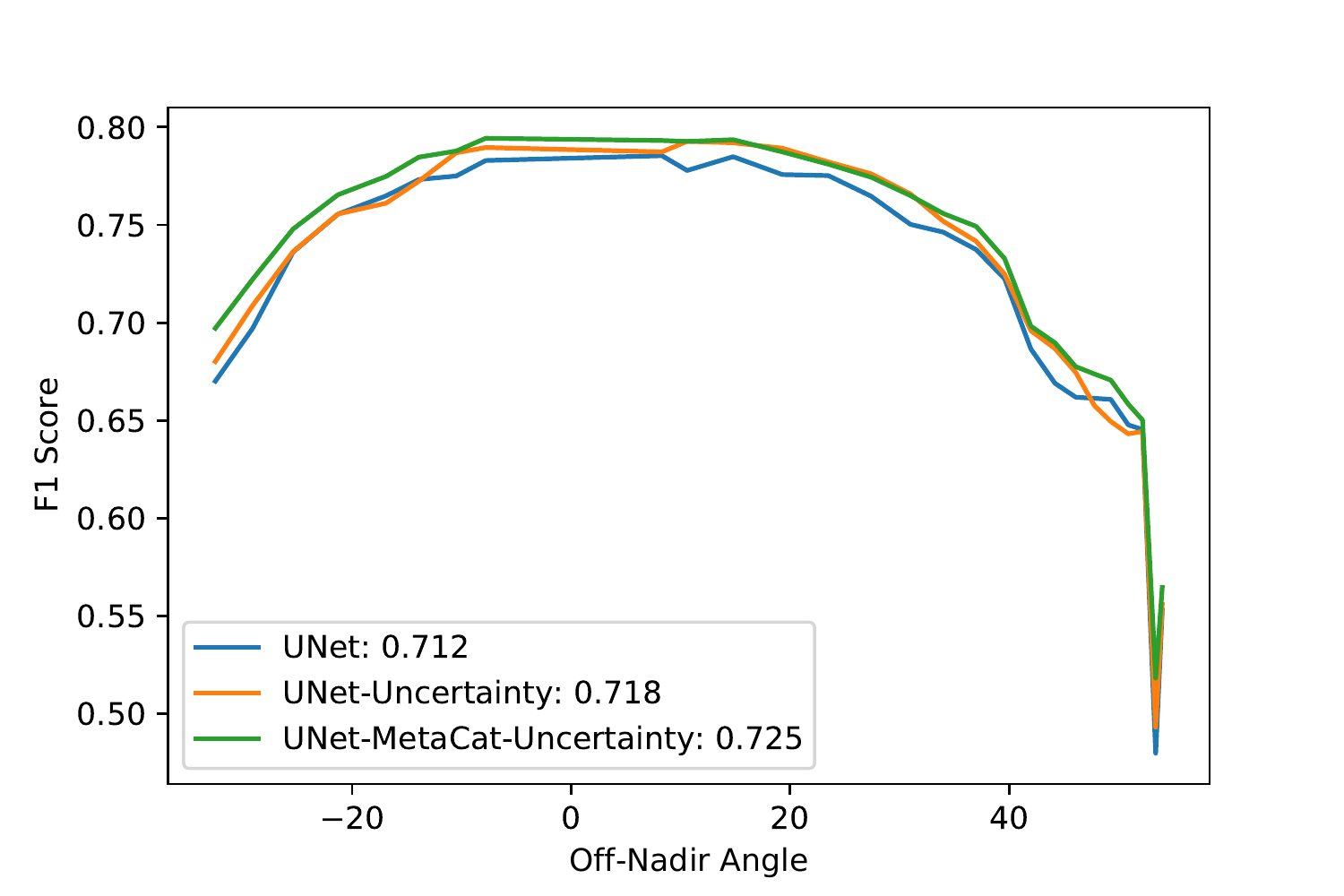}
  \caption{Testing F1 scores with different off-nadir angles. The average F1 scores of all off-nadir angles are shown in the legend.}
  \label{fig:result_f1_general}
\end{figure}

\subsection{Experimental Result and Analysis}
\label{sec:exp_result}

We start with evaluating the use of uncertainty modeling and metadata injection (we consider MetaCat first and then compare MetaCat with MetaACM later).
Figure~\ref{fig:result_f1_general} shows the F1 scores with different off-nadir angles in the testing set.
Compared with the baseline U-Net, with uncertainty modeling, there is a slight improvement across most of the off-nadir angles.
Adding the metadata injection layer can further improve the performance, especially for the cases with larger off-nadir angles ($> 40^\circ$) and negative off-nadir angles.
As mentioned in~\cite{Weir_2019}, due to the data collection process, the images with large negative off-nadir angles have very different lighting conditions and shadows.
Since most of the images are collected from positive off-nadir angles, the baseline method will suffer from unbalanced data during training.
With metadata injection and uncertainty modeling, the proposed method is able to deal with the changes of lighting and shadows.

\begin{figure*}[htbp]
  \centering
  \includegraphics[width=\linewidth]{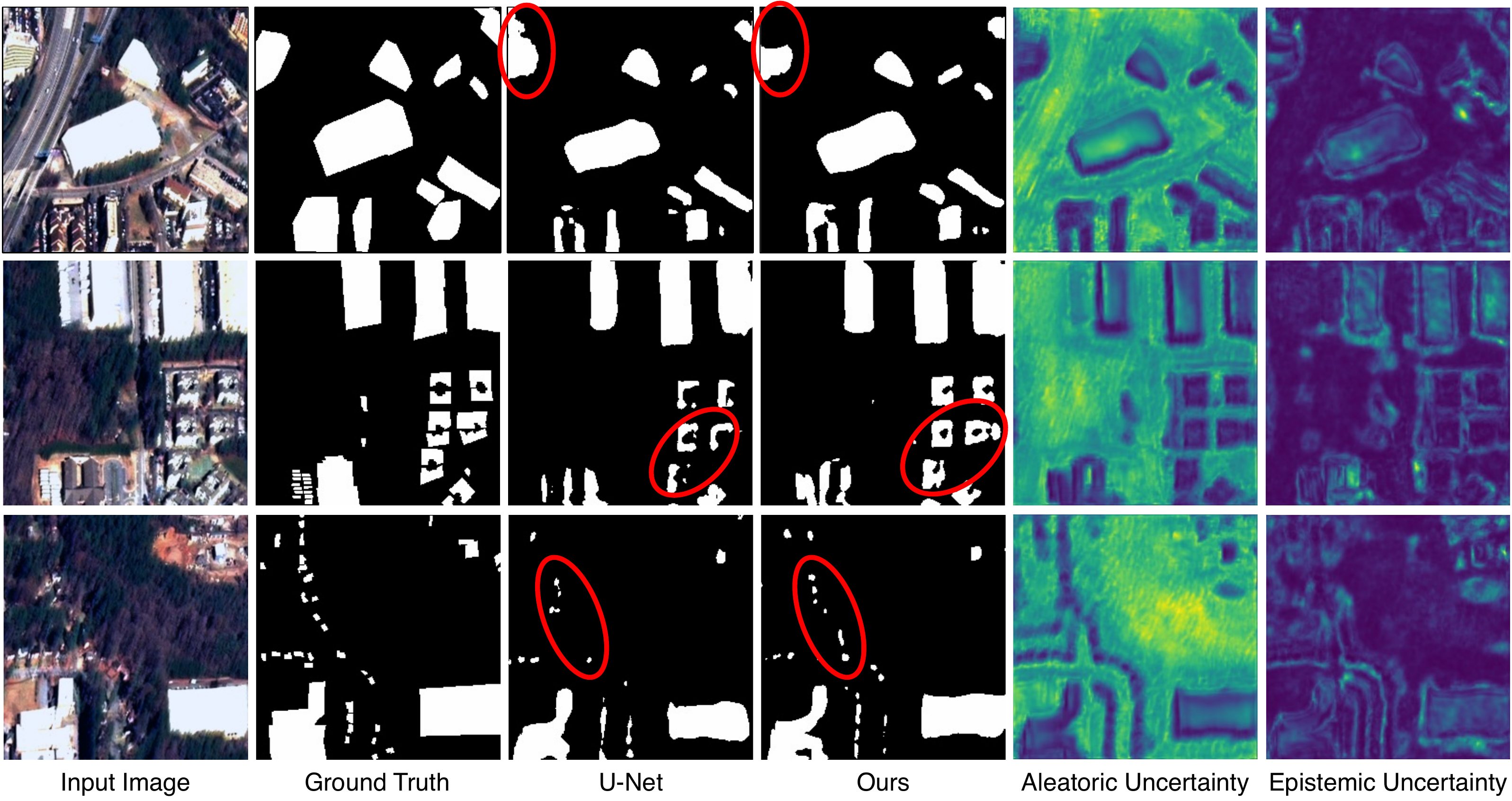}
  \caption{Result comparison of the baseline U-Net and the proposed method with uncertainty modeling and metadata injection. The input images are taken with the off-nadir angle as $54^\circ$. The red circles highlight the improvement of the proposed method compared to the baseline U-Net.}
  \label{fig:result_compare}
\end{figure*}

Figure~\ref{fig:result_compare} shows three testing examples captured from the largest off-nadir angles to visualize the improvement of the proposed method compared to the baseline U-Net.
Based on the ground truth, we can see that the proposed method is able to detect more accurate building area even under this high noise-level condition. 
For instance, in the first example, the baseline U-Net fails to differentiate the parking lot area and the building area in the top-left of the input image highlighted by the red circle).
The proposed method is able to segment the area correctly. 
From the epistemic uncertainty map, the proposed method raises higher uncertainty indicating the predictions from those class-ambiguous pixels are not reliable.
Similar examples can be found in the highlighted areas in the second and third images. 
From the aleatoric uncertainty, we can also see that the input data has higher data noise around the forest region compared to the building region.
This is due to the larger appearance variance of forests compared to buildings.
Therefore, our model will focus more on the building region during training to avoid the adverse effect of the frequent appearance changes from the forest region.
Unlike aleatoric uncertainty, epistemic uncertainty focuses more around the buildings or other man-made structures (\eg roads).
It highlights the area where the predictions are not reliable, such as the boundary of buildings, due to the image blur and noise. 
Figure~\ref{fig:teaser} shows the prediction difference of two images with same scene but different off-nadir angles.
We can see that overall, aleatoric uncertainty has a significant increase from small to large off-nadir angles due to higher noise in the input image.
Although there is less of an increase with epistemic uncertainty, the area where it highlights does get larger.
Appendix \hyperref[apx:result]{A} shows the results with more off-nadir angles for the same scene for comparison.

\begin{table*}[htbp]
  \centering
  \caption{F1 scores for the ablation study of uncertainty modeling. All of the listed experiments are based on U-Net with concatenation-based metadata injection. \textit{None} means no uncertainty modeling.}
  \label{table:result_f1_ablation}
  \begin{tabular}{ccccc}
    \toprule
    Experiment  & Nadir           & Off-Nadir       & Very Off-Nadir              & Overall \\
    \midrule
    None        & 0.7820          & \textbf{0.7450} & 0.6335                      & 0.7219  \\
    Aleatoric   & 0.7822          & 0.7448          & \tablecolor \textbf{0.6499} & \textbf{0.7275}  \\
    Epistemic   & \textbf{0.7824} & 0.7424          & \tablecolor 0.6380          & 0.7229  \\
    Both        & 0.7822          & 0.7429          & \tablecolor 0.6415          & 0.7249  \\
  \bottomrule
\end{tabular}
\end{table*}

\begin{figure*}[htbp]
  \centering
  \includegraphics[width=0.82\linewidth]{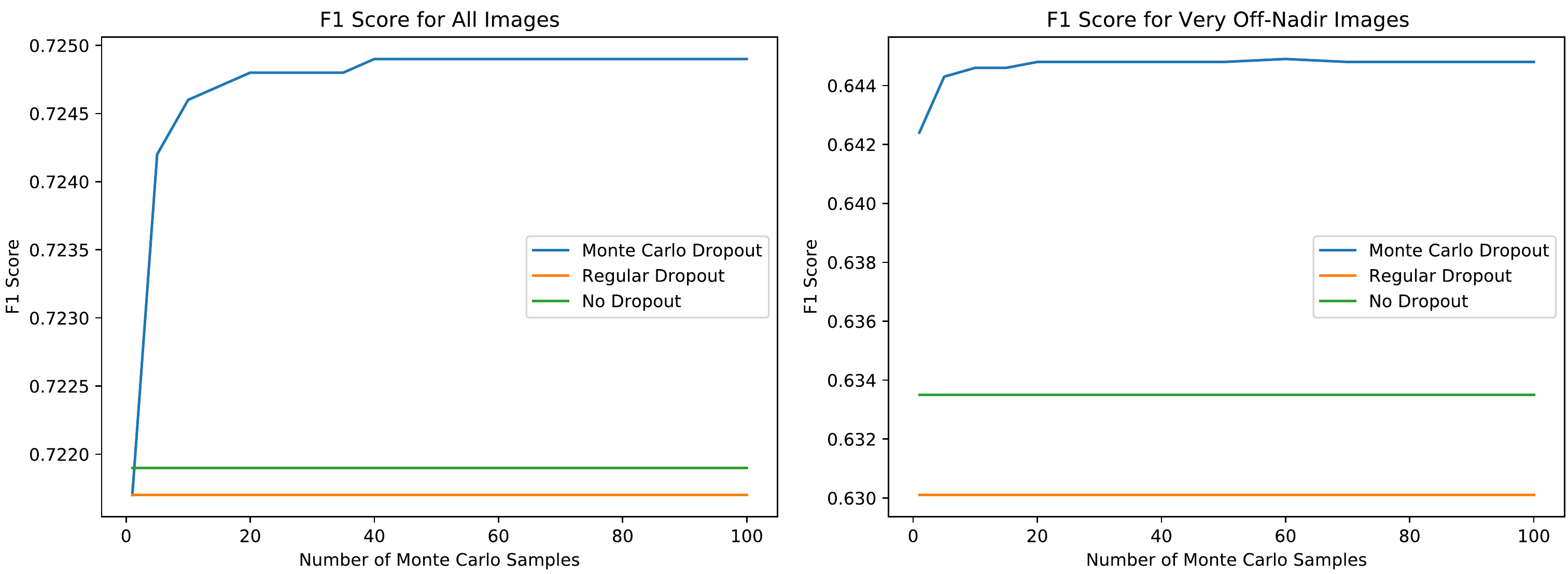}
  \caption{Ablation study of Monte Carlo dropout. F1 scores for different numbers of Monte Carlo samples are shown for all images from the validation set (left) and for the images in \textit{Very Off-Nadir} category (right).}
  \label{fig:result_f1_mc_samples}
\end{figure*}

We also provide the ablation study of uncertainty modeling to show that modeling both uncertainties does not necessarily yield the best result.
Due to the close performance of the compared experiments, we group the F1 scores with different ranges of off-nadir angle in Table~\ref{table:result_f1_ablation} to better visualize differences.
As defined in~\cite{Weir_2019}, we group the images into three categories based on the off-nadir angles $\theta$ as following:
\begin{itemize}
  \item \textit{Nadir}: $0^\circ \leq \vert \theta \vert \leq 25^\circ$;
  \item \textit{Off-Nadir}: $25^\circ < \vert \theta \vert < 40^\circ$;
  \item \textit{Very Off-Nadir}: $40^\circ \leq \vert \theta \vert < 90^\circ$.
\end{itemize}
As shown in Table~\ref{table:result_f1_ablation}, the best performance from each category are not from the experiment with both uncertainties.
Therefore, the effectiveness of uncertainty modeling could be different depending on the dataset and task.
Furthermore, as shown in the highlighted cells, for the \textit{Very Off-Nadir} category, all experiments with uncertainty modeling achieve much better performance than the method without uncertainty modeling.
This confirms that using uncertainty modeling improves the model performance when larger data noise appears.

Figure~\ref{fig:result_f1_mc_samples} shows the effectiveness of different number of samples in Monte Carlo integration obtained from our validation set. 
The \textit{Regular Dropout} experiment uses dropout as a regularization method meaning that dropout is only used during training.
The \textit{No Dropout} experiment does not use dropout for both training and testing.
From the overall F1 score plot (left) and the F1 score plot for the \textit{Very Off-Nadir} category (right), we can see that the performance stops improving when the number of samples is over $40$, which shows our choice of $50$ samples is reasonable.
Furthermore, we also show that with Monte Carlo dropout, a better result can be achieved compared to regular dropout and no dropout experiments.
Among the three experiments, regular dropout has the worst performance. 
This shows the same observation as mentioned in~\cite{Gal_2016a}, since empirically adding dropout layer in CNN tends to have a deteriorated performance.

\begin{table*}[htbp]
  \centering
  \caption{F1 scores for ACM-based and concatenation-based metadata injection. All of the listed experiments are based on U-Net with uncertainty modeling of both aleatoric and epistemic uncertainties. \textit{None} means no metadata injection.}
  \label{table:result_acm_ablation}
  \begin{tabular}{ccccc}
    \toprule
    Experiment    & Nadir           & Off-Nadir       & Very Off-Nadir  & Overall \\
    \midrule
    None          & 0.7752          & 0.7359          & 0.6347          & 0.7180  \\
    MetaCat       & \textbf{0.7822} & \textbf{0.7429} & 0.6415          & \textbf{0.7249}  \\
    MetaACM       & 0.7758          & 0.7382          & \textbf{0.6419} & 0.7197  \\
  \bottomrule
\end{tabular}
\end{table*}

We compare the ACM-based (MetaACM) and concatenation-based (MetaCat) metadata injection methods in Table~\ref{table:result_acm_ablation}.
Overall, MetaCat achieves better performance than MetaACM.
Compared with the method without metadata injection, MetaCat has significant improvement for all three off-nadir angle categories.
Although MetaACM does not have a major improvement for the lower off-nadir angle images, it achieves the best performance under the \textit{Very Off-Nadir} category.

\begin{figure*}[htbp]
  \centering
  \includegraphics[width=\linewidth]{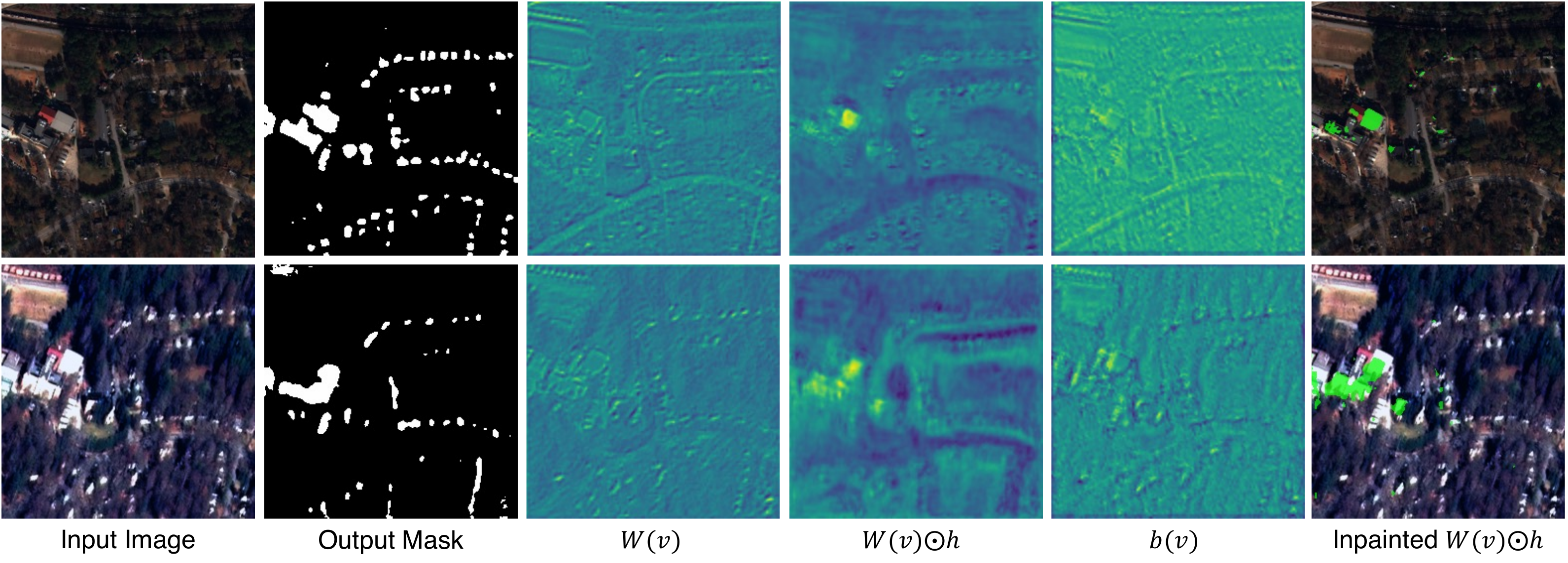}
  \caption{Illustration of ACM feature maps obtained from the last decoder layer. The inpainted results are obtained by thresholding the normalized ACM product map (green region) with threshold value as 0.5. The first row shows the case with off-nadir angle as $-7.8^\circ$. The second row shows the result of the same scene but with off-nadir angle as $54^\circ$.}
  \label{fig:result_acm}
\end{figure*}

Figure~\ref{fig:result_acm} shows the ACM feature maps obtained from the last decoder layer. 
Following~\cite{Li_2020}, the visualization of these feature maps is obtained by computing the average along the channel dimension. 
The results from the fourth column show the $\mathbf{h} \odot W(\mathbf{v})$ map based on Equation~\ref{eq:acm}.
As we discussed in Section~\ref{sec:metaacm}, this feature map should highlight the metadata-relevant information, since it directly interacts with the metadata features (or the previous decoder features). 
The first row in Figure~\ref{fig:result_acm} shows the case with small off-nadir angle.
Its $\mathbf{h} \odot W(\mathbf{v})$ map mainly addresses the entire building area, according to the inpainted result from the last column.
However, when dealing with a large off-nadir angle, the $\mathbf{h} \odot W(\mathbf{v})$ map highlights the lower side of building area, as shown in the second row of Figure~\ref{fig:result_acm}.
With larger off-nadir angle, building facade becomes visible which increases the building area compared to the case with small off-nadir angle.  
ACM highlights the building facades (lower side of the building area) to improve the prediction on those regions. 
This confirms our observation in Table~\ref{table:result_acm_ablation} that MetaACM is able to significantly improve the performance of the \textit{Very Off-Nadir} category. 

\begin{figure*}[htbp]
  \centering
  \includegraphics[width=\linewidth]{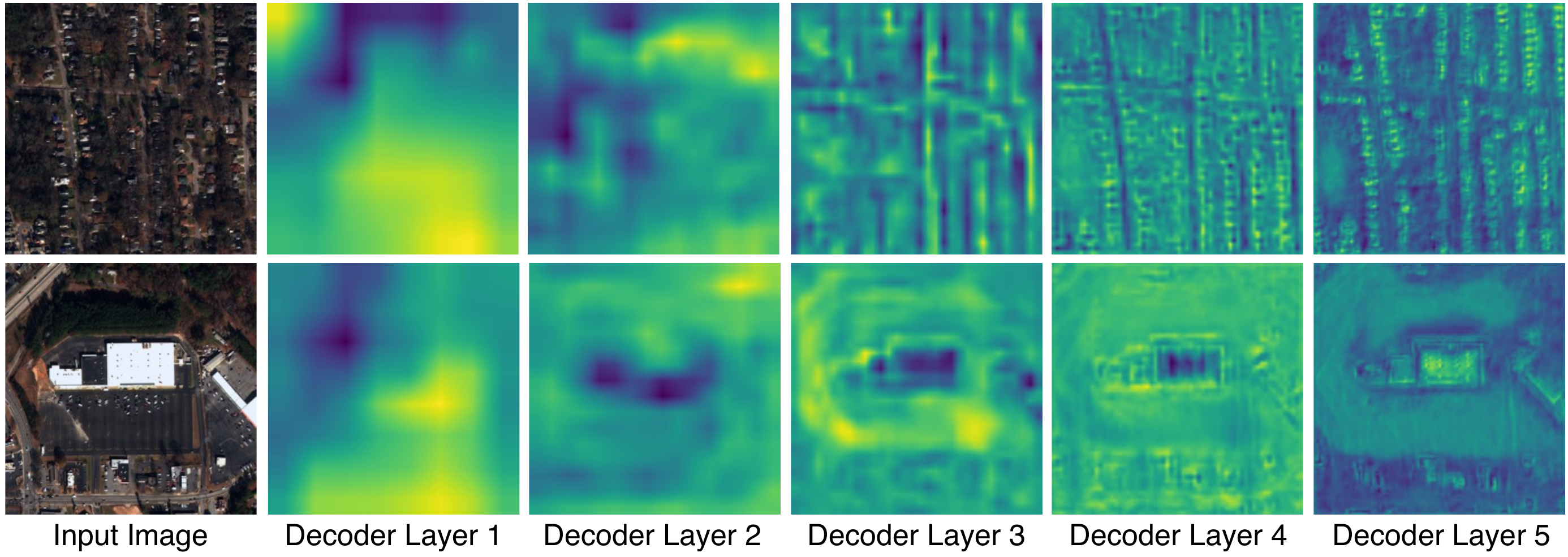}
  \caption{Resized ACM $W(v)\odot h$ map for different decoder layers. The resolution of ACM map from decoder layer 1 is $8\times8$ and increases with the factor of 2 after each decoder layer.}
  \label{fig:result_acm_layers}
\end{figure*}

Figure~\ref{fig:result_acm_layers} shows the ACM $\mathbf{h} \odot W(\mathbf{v})$ map from different decoder layers.
We can see that the feature maps from different decoder layers address different part of the image.
The design of our MetaACM enables the model to locate different areas for different feature resolutions.
This is important for metadata injection, since if we only modify the image features using metadata features in the lowest resolution (\eg MetaCat), these modifications will affect a large area in the final full-resolution result.
For example, in our case, the bottleneck layer has the resolution of $8 \times 8$ and the final result has the resolution of $256 \times 256$. 
If we only consider the effect of upsampling operators (without considering the change of receptive field caused by convolution), any modifications of the features from the bottleneck layer will affect at least $32 \times 32$ area in the final result.
These modifications are not accurate enough for the buildings that are much smaller than $32 \times 32$ pixels.
Therefore, injecting metadata features for the image features with different resolutions is important for the refinement of small buildings. 
To show the effectiveness of the proposed ACM-based metadata injection method, we also evaluate it on a different backbone model, U$^2$-Net~\cite{Qin_2020}.
Please check Appendix \hyperref[apx:u2net]{B} for more information on the experiments.

\section{Conclusion}

In this paper, we propose a method that can provide accurate building segmentation despite the data noise that is caused by large off-nadir angles.
Both aleatoric uncertainty and epistemic uncertainty are modeled by our method to enable our model to learn from noisy training data.
Based on the level of predicted uncertainty, the proposed method learns to ignore the area with larger uncertainty and focus on the area with less uncertainty.
Satellite image metadata is also considered to further improve the performance.
We propose concatenation-based and ACM-based metadata injection methods to effectively use metadata for the building segmentation task.
By conducting the experimental analysis and ablation study, we show that the proposed method is able to achieve a clear improvement compared to the baseline method, especially for the noisy images taken from large off-nadir angles.

\section*{Acknowledgments} 

This material is based on research sponsored by Lockheed Martin Space. 
The views and conclusions contained herein are those of the authors and should not be interpreted as necessarily representing the official policies or endorsements, either expressed or implied of Lockheed Martin Space.

{\small
\bibliographystyle{ieee_fullname}
\bibliography{reference}
}

\sectionprelude

\section*{Appendix A: Result Comparison with Different Off-Nadir Angles} 
\label{apx:result}

\begin{figure*}[htbp]
  \centering
  \includegraphics[width=\linewidth]{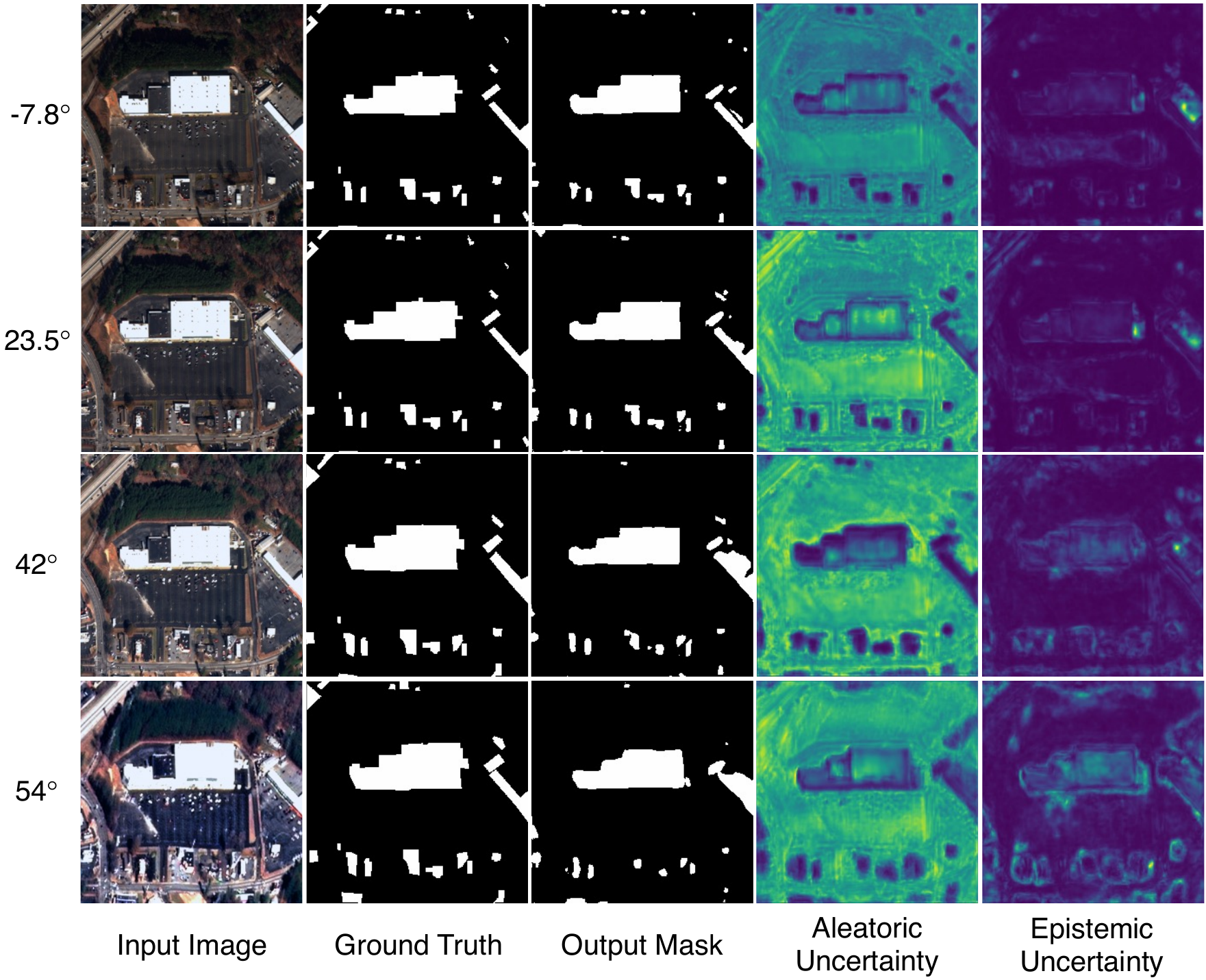}
  \caption{Results of the proposed method for the images taken from different off-nadir angles. The results are obtained from the model with uncertainty modeling and concatenation-based metadata injection.}
  \label{fig:result_uncertainty_more}
\end{figure*}

Figure~\ref{fig:result_uncertainty_more} shows the proposed method with uncertainty modeling and concatenation-based metadata injection.
All four input images are taken from the same scene but with different off-nadir angles from $-7.8^\circ$ to $54^\circ$.
From the aleatoric uncertainty and epistemic uncertainty maps, we can see that larger off-nadir angle leads to higher uncertainty. 
More specifically, as off-nadir angle gets larger, the aleatoric uncertainty increases for the non-building regions, like forests and parking lots, since these areas are usually contains larger appearance variance. 
By raising higher uncertainty around these regions, our method can ignore those regions and address more on the building areas. 
The epistemic uncertainty mainly highlights the building boundaries, since the predictions from these areas are not reliable compared to other regions. 
As the off-nadir angle gets larger, the highlighted building boundaries get thicker, because these images get noisier and blurrier.
Therefore, with uncertainty modeling, the proposed method is able to learn from the image area with less uncertainty without getting deteriorated by the noisy areas. 

\section*{Appendix B: U$^2$-Net Result} 
\label{apx:u2net}

\begin{figure*}[htbp]
  \centering
  \includegraphics[width=0.7\linewidth]{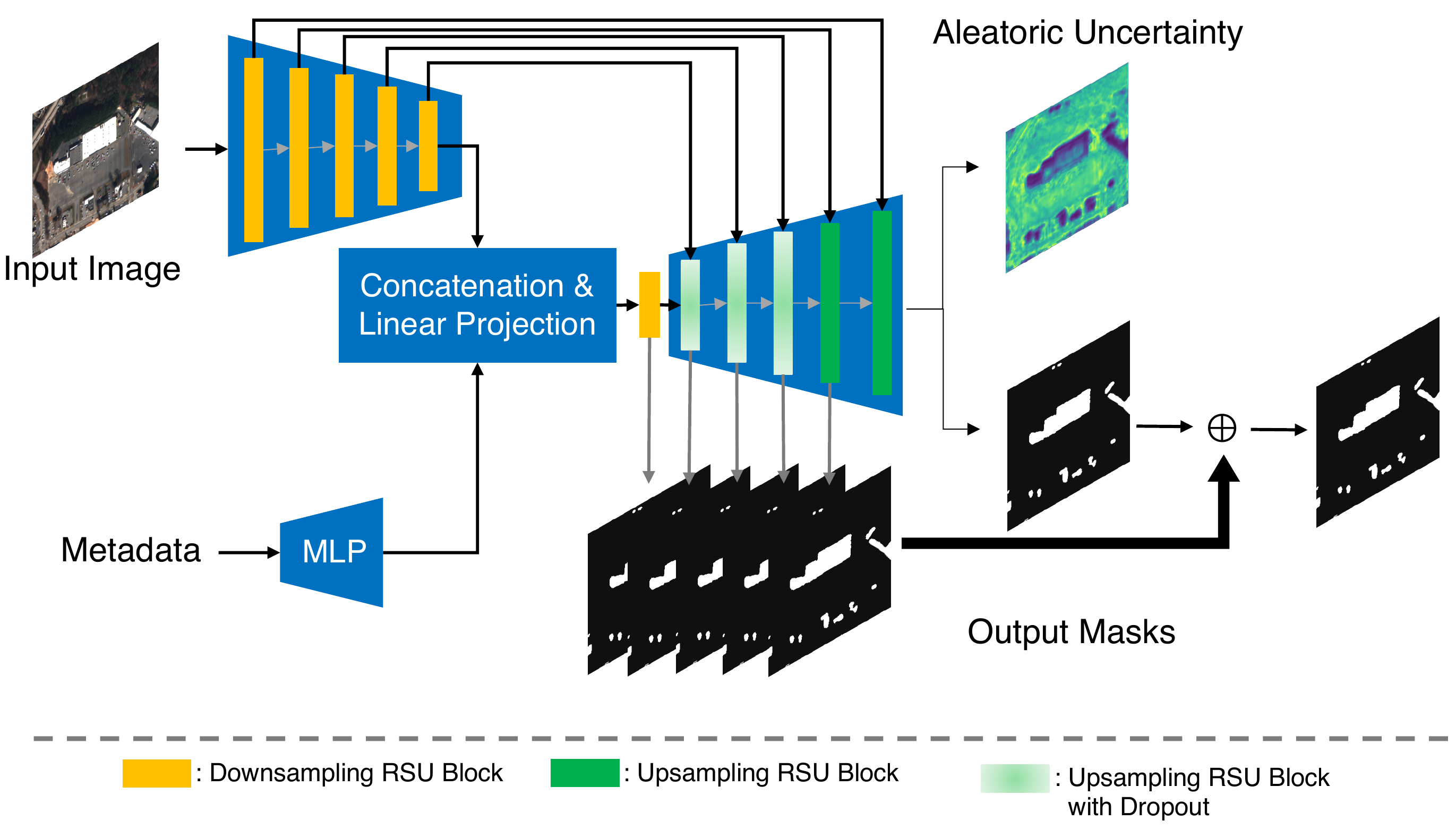}
  \caption{The block diagram of the proposed U$^2$-Net~\cite{Qin_2020} with uncertainty modeling and concatenation-based metadata injection.}
  \label{fig:u2net-cat}
\end{figure*}

\begin{table*}[htbp]
  \centering
  \caption{F1 scores of U$^2$-Net with uncertainty modeling and metadata injection. \textit{None} means no metadata injection and no uncertainty modeling. The experiments with \textit{Uncertainty} use both aleatoric and epistemic uncertainties.}
  \label{table:result_u2net}
  \begin{tabular}{ccccc}
    \toprule
    Experiment                  & Nadir           & Off-Nadir       & Very Off-Nadir  & Overall \\
    \midrule
    None                        & 0.8019          & 0.7447          & 0.6185          & 0.7259  \\
    Uncertainty                 & 0.8081          & 0.7588          & 0.6305          & 0.7356  \\
    Uncertainty + MetaCat       & 0.8080          & 0.7580          & 0.6137          & 0.7304  \\
    Uncertainty + MetaACM       & \textbf{0.8163} & \textbf{0.7700} & \textbf{0.6348} & \textbf{0.7426}  \\
  \bottomrule
\end{tabular}
\end{table*}

The proposed uncertainty modeling and metadata injection methods can be extended to other backbone models. 
As shown in Figure~\ref{fig:u2net-cat}, we can apply the proposed methods for U$^2$-Net~\cite{Qin_2020}, which is a modified version of the original U-Net. 
It is able to utilize a two-level nested U-structure to enlarge the receptive field in each encoder/decoder block.
Moreover, deep supervision~\cite{Xie_2017} (\ie output multiple masks for different decoder blocks) is used to enforce the integration of multi-level deep features to further improve the performance.
Please check the original paper~\cite{Qin_2020} for the detailed design of U$^2$-Net. 
Similar to the U-Net backbone, we use the epistemic uncertainty modeling (\ie Monte Carlo dropout layers) in the first three decoder blocks in U$^2$-Net.
Then we split the final layer into two branches to learn the aleatoric uncertainty map.
Figure~\ref{fig:u2net-cat} shows the model with concatenation-based metadata injection.
The ACM-based metadata injection method can be obtained in a manner similar to the block diagram shown in Figure~\ref{fig:method_acm}.

Table~\ref{table:result_u2net} shows the results of U$^2$-Net with uncertainty modeling and the different metadata injection approaches. 
Using the proposed uncertainty modeling and metadata injection methods can improve the original U$^2$-Net model, especially for the cases with large off-nadir angles (except the \textit{Very Off-Nadir} case from the concatenation-based metadata injection experiment). 
The experiment with both uncertainty modeling and the ACM-based metadata injection method achieves the best performance.
It achieves the best performance for all off-nadir angle cases, which confirms the benefit of using the multi-level features in metadata injection.
Therefore, from the aforementioned experiments, we show that the proposed uncertainty modeling and metadata injection methods can improve the performance of both U-Net and U$^2$-Net.

\end{document}